\documentclass{ifacconf}

\usepackage{graphicx}      
\usepackage[square,numbers]{natbib}        
\usepackage{amssymb}
\usepackage{mathtools}
\usepackage{amsmath,bm}
\usepackage{amsmath}
\usepackage{multirow}
\usepackage{xcolor}
\usepackage{subcaption}
\usepackage{array}

\begin{document}
\begin{frontmatter}

\title{Towards Safety Assured End-to-End Vision-Based Control for Autonomous Racing} 




\author{Dvij Kalaria$^{1}$, Qin Lin$^{2}$, and John M. Dolan$^{1}$}
\thanks{Dvij Kalaria and John M. Dolan are with the Robotics Institute, Carnegie Mellon University {\tt\small \{dkalaria,jdolan\}@andrew.cmu.edu}}%
\thanks{Qin Lin is with the EECS Department, Cleveland State University {\tt\small q.lin80@csuohio.edu}}

\begin{abstract}                
Autonomous car racing is a challenging task, as it requires precise applications of control while the vehicle is operating at cornering speeds. Traditional autonomous pipelines require accurate pre-mapping, localization, and planning which make the task computationally expensive and environment-dependent. Recent works propose use of imitation and reinforcement learning to train end-to-end deep neural networks and have shown promising results for high-speed racing. However, the end-to-end models may be dangerous to be deployed on real systems, as the neural networks are treated as black-box models devoid of any provable safety guarantees. 
In this work we propose a decoupled approach where an optimal end-to-end controller and a state prediction end-to-end model are learned together, and the predicted state of the vehicle is used to formulate a control barrier function for safeguarding the vehicle to stay within lane boundaries. We validate our algorithm both on a high-fidelity Carla driving simulator and a 1/10-scale RC car on a real track. The evaluation results suggest that using an explicit safety controller helps to learn the task safely with fewer iterations and makes it possible to safely navigate the vehicle on the track along the more challenging racing line. 
\end{abstract}


\end{frontmatter}

\section{Introduction}

High-speed autonomous racing presents unique
challenges that have gained recent attention after significant progress in urban autonomous driving. The following items make autonomous racing more challenging than traditional urban driving: 1) The vehicle operates near its handling limits in a highly non-linear regime \citep{Liniger2017OptimizationbasedAR}. 2) It has to avoid collisions in a rapidly changing environment \citep{amz_driverless,kalaria2022delay}. 3) It needs to have very low response time given its high-speed nature. The classical pipeline splits the task into various traditional components, i.e.,  pre-mapping, localization, planning, and control \citep{amz_driverless,osti_10296575}. However, such a pipeline requires significant onboard computation resources. Also, these methods are prone to environment changes and often require pre-mapping of the race track to operate safely and efficiently \citep{osti_10296575}. 

Recent advances in deep learning have brought new approaches to handing autonomous driving/racing based on an end-to-end solution. Based on the powerful
representation abilities of deep neural networks, end-to-end methods directly take as input only the raw high-dimensional sensory data (e.g., RGB images and/or Lidar pointclouds) and output low-level control commands (i.e., steering and throttle). Most of these DNNs representing end-to-end control policies are trained with imitation learning (IL),
which can efficiently use expert driver demonstrations (i.e., observation-action data pairs) to train efficeient models, but suffers from the distribution mismatch problem (a.k.a. covariate shift). Specifically, IL assumes the dataset is independent and identically distributed (i.i.d.) and trains networks to predict actions offline that do not affect future states. However, driving a car is inherently a sequential task, and the actions predicted by the network indeed affect the future. This phenomenon breaks the i.i.d. assumption and can lead to a distribution shift between training and testing. Thus, the policy network may make mistakes on unfamiliar state distributions after deployment. To address this issue, recent approaches use DAgger \citep{Ross2011ARO} or online supervision \citep{Pan2017AgileOA}, but they bring further problems of using human demonstrations for accurately labeling the data. To alleviate this, some approaches use supervised control, which takes the complete accurate state of the vehicle as input to output control commands to train the end-to-end model, which then learns to autonomously navigate the track.

However, these approaches have the following limitations: 1) They treat the end-to-end trained model as a black box, the performance is not robust to environmental changes, and they do not provide any safety guarantees. 2) DNNs require a lot of training data whose collection may be inconvenient and practically difficult on a real race track. 3) The end-to-end network does not explicitly take vehicle dynamics into account for its training or for ensuring safety. Hence, we require a \textbf{data-efficient approach which can safeguard the end-to-end controller to run on a wide range of environments}. To this end, we propose the use of two different deep neural networks which are learned to output optimal control and the state of the vehicle respectively coupled with use of an IMU sensor and wheel encoder sensors to get the dynamic state of the vehicle. This predicted dynamic state is used to formulate a probabilistic control barrier function (CBF) which makes use of the vehicle dynamics to guarantee safety of the vehicle. The epistemic uncertainty in the DNN predictions is estimated using a dropout-based approach \citep{Gal2016DropoutAA} to account for uncertainty in unknown observations. Another common drawback in previous works is that most of them \citep{Wadekar2021TowardsED, Pan2017AgileOA, Cai2021VisionBasedAC} aim to train the end-to-end controller on the center line, which we believe makes the task relatively easy, as the there is ample scope for the vehicle to recover from moving out of the lane boundaries. We, however, aim to teach the end-to-end controller to run on the racing line at operating limits, which turns out to be more challenging, as the racing line is mostly located near the lane boundaries, leaving less scope for the end-to-end controller to make mistakes. Hence, we decouple the optimal end-to-end controller, which aims to track the racing line in order to achieve minimum lap time, and the safety controller, which tries to ensure vehicle safety.

Finally, we demonstrate our framework running on a 1/10-sized RC car with limited computation power and sensors. To summarize, we make the following contributions:

1) We propose an uncertainty-aware framework which decouples the safety aspect of the controller and the optimality of the controller. Probabilistic CBFs are used to explicitly impose probabilistic safety constraints on the end-to-end controller.

2) We run the racing car on the racing line at near to cornering speeds instead of the center line with constant speed, unlike most recent works.

3) We propose use of other dynamic model inputs like lateral velocity and angular velocity of the vehicle as input to the end-to-end DNN architecture.

4) We validate the proposed method both in a realistic driving simulation (Carla) and on a real-world 1/10-scale RC car for autonomous racing, showing that it outperforms previous traditional implicit end-to-end learning approaches.

To the best of our knowledge, this is the first work proposing use of CBFs on predicted state from a learned model to safeguard end-to-end models. The rest of the paper is organized as follows: Sec. \ref{sec:il} discusses the proposed imitation learning framework we use. Sec. \ref{sec:safe_cont} describes the safety controller design we use to safeguard the end-to-end optimal controller. Sec. \ref{sec:tr_cycle} describes the overall training cycle and finally Sec. \ref{sec:conclusion} concludes the paper.

\section{Imitation learning for autonomous racing} \label{sec:il}

\subsection{Problem definition}

We first mathematically formulate the autonomous racing task. Let us consider a discrete-time continuous-valued optimal control problem. Let $\mathcal{S}$, $\mathcal{A}$, and $\mathcal{O}$ denote the state, action and observation spaces respectively. In our setting, we assume that the observations of the system are the RGB camera image from the front camera and the longitudinal, lateral, and angular velocities from the wheel encoders and the IMU sensor. The action space consists of the continuous-valued steering and throttle commands.

The goal is to find a policy $\pi : \mathcal{O} -> \mathcal{A}$ which minimizes the cumulative cost over a finite horizon of length $T$ denoting the whole episode as follows:

\begin{equation} \label{problem_def}
    \mathop{\min}\limits_{\pi} \quad \Sigma_{i=0..T-1} C_i 
\end{equation}

in which $s_t \in \mathcal{S}$, $a_t \in \mathcal{A}$ and $\rho_\pi$ is the distribution of the trajectory $(s_0,o_0,a_0,...,a_{T-1})$ under policy $\pi$. For the racing scenario, we usually choose function $C$ to encourage that the car stays within the track and completes the lap in minimum time or makes maximum progress along the track safely. Our goal is to learn optimal policy $\pi$ such that taking $a_t=\pi(o_t)$ sequentially leads to minimizing the cumulative cost $J(\pi)$.

\subsection{Imitation learning}

Directly optimizing Eq. \ref{problem_def} is challenging for high-speed autonomous racing. Also, model-free RL techniques are not very effective in our case, which involves a physical robot, as they are quite sample-inefficient and thus would be quite inconvenient. Using model-based RL may require fewer samples, but it can lead to unstable results, as it is difficult to learn from the RGB image and speed inputs as the state space due to its high dimension. 

Considering these limitations, we propose to use Imitation Learning. Given an expert policy $\pi^*$, we would like to match our policy $\pi$ as closely to $\pi^*$ as possible. For training the policy $\pi$, we use an expert controller to get the policy $\pi^*$. We assume access to additional sensors like Lidar, GPS etc. and computation for deriving the expert controller i.e., the expert may be a computationally intensive optimal controller that relies on GPS and Lidar for accurate state estimation. Our goal is to learn the policy $\pi$ which performs similarly to $\pi^*$, but only takes low-cost camera sensors and IMU sensors as input with limited computation power for it to run during test time. Put formally, the state $s^*$ used by the expert consists of the computed accurate position, orientation of the vehicle on the track using GPS and Lidar, speed inputs from IMU, and wheel encoders; while the state $s$ used by policy $\pi$ consists of just the camera RGB image(s) and speed inputs. Here, we assume that the localized position and heading angle should map one-to-one to the camera observation, which means that we have a unique image observation for a given state of the vehicle, i.e., there should exist a function $F$ that performs a one-to-one mapping $s^*->s$. This would be reasonably possible for a camera attached on the car in a fixed frame of reference w.r.t. the car's body.   

\subsection{DAgger} 

To train the policy $\pi$, we use the DAgger algorithm \citep{Ross2011ARO}. We start with an empty dataset, $D=\phi$, then we run the vehicle with an optimal controller following policy $\pi^*$ described later. The optimal policy makes use of high computation resources and extensive sensors for its training. Also, a camera is attached to the vehicle whose observations are used for creating the dataset for training the policy $\pi$. $\pi_i$ is the trained policy obtained after $i^{th}$ iteration. The collected data are used to train the policy $\pi_i$ described by a DNN described later in Section \ref{sec:arch}. The trained policy $\pi_i$ is used to run the race car on the track and the collected data are added to the dataset $D$ for re-training for the next iteration. 


\subsection{Network architecture} \label{sec:arch}

\begin{figure}
    \centering
    \includegraphics[width=0.48\textwidth]{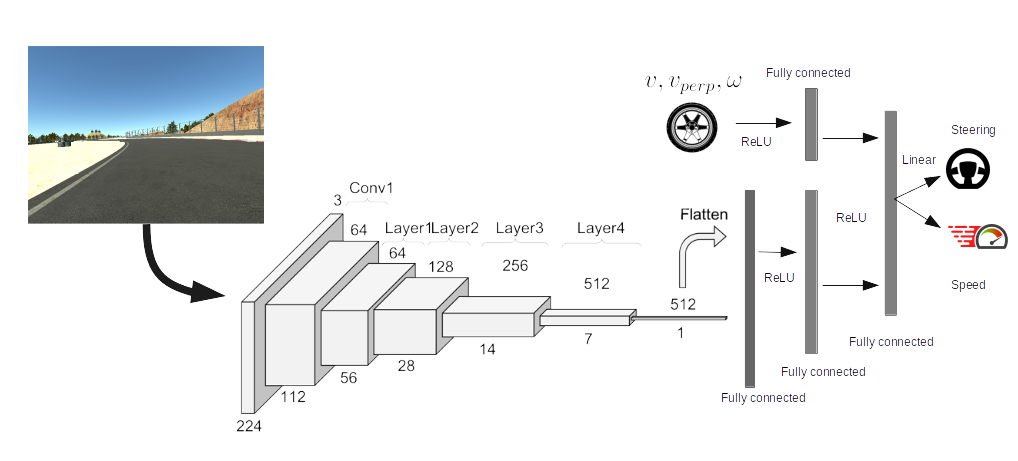}
    \caption{Control prediction model architecture}
    \label{fig:model_arch}
\end{figure}

For the policy $\pi$, we use a DNN with network architecture similar to \cite{Pan2017AgileOA}, as shown in Fig. \ref{fig:model_arch}. It consists of a modified ResNet18 architecture with removal of the last classification layer being fed by the camera RGB image as input to get a one-dimensional feature vector. The lateral and longitudinal speeds and angular velocity from the IMU sensor, and the wheel encoder are appended to this feature vector, and the resultant feature vector is passed through a fully connected layer to get the steering angle and target speed as outputs. The target speed is achieved though a simple PID longitudinal controller that outputs the required throttle.    

\subsection{Expert controller} \label{expert_controller}

The expert controller using MPC \citep{amz_driverless} basically minimizes the cost $C$ formally defined as:

\begin{equation} \label{objective function}
\begin{split}
    \mathop{\min}\limits_{U}&\quad\sum_{k=0}^{N-1} (\Bar{\bm{x}}_k-{\bm{x}_{ref,k}})^T Q_k (\Bar{\bm{x}}_k-{\bm{x}_{ref,k}})+\Bar{\bm{u}}_k^T R_k \Bar{\bm{u}}_k\\&+(\Bar{\bm{x}}_N - {\bm{x}_{ref,N}})^T Q_N (\Bar{\bm{x}}_N - {\bm{x}_{ref,N}})\\
    s.t. &\quad \Bar{\bm{x}}_{k+1} = A \Bar{\bm{x}}_k + B \Bar{\bm{u}}_k\\
    &\quad \bm{x}_{0} = \bm{x}_{start}\\
    &\quad \bm{x_{k+1}} = A_k \bm{x_k} + B_k \bm{u_k} + C_k
\end{split}
\end{equation}

where, $A_k$, $B_k$, $c_k$ are the state, control matrices respectively obtained by linearizing the vehicle dynamics at the current state. $\bm{x}_{ref,k}$ are the reference states obtained to follow the racing line reference described in Section \ref{racing_line}. The vehicle models $F$ are further described in section \ref{sec:veh_models}.

\subsection{Optimal racing line} \label{racing_line}

During a race, professional drivers follow a racing line while using specific maneuvers that allow them to use the limits of the car’s tire forces. This path can be used as a reference by the autonomous racing motion planner to effectively follow time-optimal trajectories while avoiding collision, among other objectives. The racing line is essentially a minimum-time path or a minimum-curvature path. They are similar but the minimum-curvature path additionally allows the highest cornering speeds given the maximum legitimate lateral acceleration \citep{doi:10.1080/00423114.2019.1631455}. 

In our work, we calculate the minimum-curvature optimal line, which is close to the optimal racing line as proposed by \cite{doi:10.1080/00423114.2019.1631455}. The race track information is input by a sequence of tuples ($x_i$,$y_i$,$w_i$), $i \in \{0,...,N-1\}$, where ($x_i$,$y_i$) denotes the coordinate of the center location and $w_i$ denotes the lane width at the $i^{th}$ point, vehicle width $w_{veh}$. The output trajectory consists of a tuple of seven variables: coordinates $x$ and $y$, curvilinear longitudinal displacement $s$, longitudinal velocity $v_x$, acceleration $a_x$, heading angle $\psi$, and curvature $\kappa$. The trajectory is obtained by minimizing the following cost:
\begin{equation} \label{opt_racing_line_eqn}
\begin{split}
    & \mathop{\min}\limits_{\alpha_1...\alpha_N} \quad \sum_{n=0}^{N-1} \kappa_i^2(n)\\
    \text{s.t.} & \quad \alpha_i \in \left[ -w_i+\frac{w_{veh}}{2},w_i-\frac{w_{veh}}{2} \right]\\
\end{split}
\end{equation}
where $\alpha_i$ is the lateral displacement at the $i^{th}$ position. To generate a velocity profile, the vehicle's velocity-dependent longitudinal and lateral acceleration limits are required  \citep{doi:10.1080/00423114.2019.1631455}. 

\section{Safety controller} \label{sec:safe_cont}

In these section, we describe the design of the safety controller using CBF with predicted state as input. 

\subsection{Control Barrier Functions (CBFs)}

CBFs ensure safety by rendering a forward-invariant safe set. We define a continuous and differentiable safety function 
$h(x): \mathcal{X} \xrightarrow{} \mathbb{R}$. The 
superlevel set $\mathcal{C} \in \mathbb{R}^n$ can be named as a safe set. Let the set $\mathcal{C}$ obey

\begin{align}
    \mathcal{C} = \{\mathbf{x} \in \mathcal{X}: h(\mathbf{x})\geq 0 \} \\
    \label{eq:C}
    \partial \mathcal{C} = \{\mathbf{x} \in \mathcal{X}: h(\mathbf{x})=0 \} \\
    \text{Int}( \mathcal{C}) = \{\mathbf{x} \in \mathcal{X}: h(\mathbf{x})>0 \}.
\end{align}

A control affine system has the form $\dot{\mathbf{x}} = f(\mathbf{x})+g(x)\mathbf{u}$, such that
\begin{equation}
    \exists u \quad \text{s.t.} \quad \dot{h}(\mathbf{x}) \geq -\kappa_h (h(\mathbf{x}))
\end{equation}
where $\kappa_h \in \mathcal{K}$ is particularly chosen as $\kappa_h(a) = \gamma a$ for a constant $\gamma > 0$. The time derivative of $h$ is expressed as

\begin{equation}
    \dot{h}(\mathbf{x}) = L_fh(\mathbf{x}) + L_g h(\mathbf{x}) \mathbf{u}.
    \label{eq:hdot}
\end{equation}
where $L_f h(\mathbf{x})$ and $L_g h(\mathbf{x})$ represent the Lie derivatives of the system denoted as $\nabla h(\mathbf{x}) f(\mathbf{x})$ and $\nabla h(\mathbf{x}) g(\mathbf{x})$, respectively. The safety constraint for a CBF is that there exists a $\gamma >0$ such that

\begin{equation}
    \underset{\mathbf{u} \in \mathcal{U}}{\inf}(L_f h(\mathbf{x}) + L_g h(\mathbf{x}) \mathbf{u} ) \geq -\gamma (h(\mathbf{x}))
    \label{cbf}
\end{equation}
for all $\mathbf{x} \in \mathcal{X}$. The solution $\mathbf{u}$ assures that the set $\mathcal{C}$ is a forward invariant, i.e.,  $x(t \xrightarrow{} \infty) \in \mathcal{C}$. In a practical collision avoidance task, the safety function can be designed as the relative distance between the ego system and a dynamic obstacle.

We consider lane violation constraints for both left and right lane boundaries. We assume constant lane widths $L$ throughout the track. We consider using 3-order order CBF as we have a system with a relative degree of 3. For derivations of expression for Higher Order CBF's (HOCBF), readers are referred to \citep{Xiao2019ControlBF}. We also constraint the relative angle with respect to the frenet frame, $\theta$ within $-\theta_{max}$ and $\theta_{max}$. This helps to stabilize the vehicle at high speeds. To impose this constraint, we use $2^{nd}$ order CBF. We take $\alpha_1 = \alpha_2 = \alpha_3 = \lambda$ as per the notations in \citep{Xiao2019ControlBF}. 

\begin{equation} \label{lane_cbf}
\small
\begin{split}
    &h_{left}(X,U) = \frac{L}{2} - x, h_{right}(X,U) = \frac{L}{2} + x\\
    &\dot{x} = v_{perp} \cos(\theta) + v \sin(\theta)\\
    &\dot{v}_{perp} = \frac{-2 (C_f+C_r)}{m v} v_{perp}\\
    &\dot{\theta} = \omega - v c \\
    &\ddot{x} = \dot{v}_{perp} \cos(\theta) + a_x \sin(\theta)+(\omega - v c) (-v_{perp}\sin(\theta)+v\cos(\theta))\\
    &\dot{\omega} = \frac{-2 (l_f^2 C_f + l_r^2 C_r)}{Iz v} \omega+\left(2 \frac{l_f C_f}{Iz}\right) \delta\\
    &\ddot{v}_{perp} = \frac{-2 (C_f+C_r)}{(m v)} \dot{v}_{perp} + \frac{2 (C_f+C_r)}{(m v^2)} a_x\\
    &\dddot{x} = \ddot{v}_{perp} \cos(\theta) + (\dot{\omega}-a_x c) (-v_{perp} \sin(\theta)+v\cos(\theta)) - (\omega - v c)^2 \dot{x} \\
    &\text{3-order CBF constraint for left boundary is:}\\
    &\dddot{h}_{left}(X,U) + 3 \lambda \ddot{h}_{left}(X,U) + 3 \lambda^2 \dot{h}_{left}(X,U) + \lambda^3 h_{left}(X,U) \ge 0 \\
    &-\dddot{x} - 3 \lambda \ddot{x} - 3 \lambda^2 \dot{x} + \lambda^3 \left(\frac{L}{2} - x\right) \ge 0 \\
    &\text{3-order CBF constraint for right boundary is:}\\
    &\dddot{h}_{right}(X,U) + 3 \lambda \ddot{h}_{right}(X,U) + 3 \lambda^2 \dot{h}_{right}(X,U) \\
    &+ \lambda^3 h_{right}(X,U) \ge 0 \\
    &\dddot{x} + 3 \lambda \ddot{x} + 3 \lambda^2 \dot{x} + \lambda^3 \left (\frac{L}{2} + x \right) \ge 0 \\
    &\text{Finally, }\\
    &h_{\theta,right} = \theta + \theta_{max}, h_{\theta,left} = -\theta + \theta_{max} \\
    &\text{2-order CBF constraints for relative angle restriction are:}\\
    &\ddot{h}_{\theta,right}(X,U) + 2 \lambda \dot{h}_{\theta,right}(X,U) + \lambda^2 h_{\theta,right}(X,U) \ge 0\\
    &\text{Or } \dot{\omega} - a_x c + 2 \lambda (\omega - v c) + \lambda^2 (\theta + \theta_{max}) \ge 0 \\
    &\text{And } \ddot{h}_{\theta,left}(X,U) + 2 \lambda \dot{h}_{\theta,left}(X,U) + \lambda^2 h_{\theta,left}(X,U) \ge 0\\
    &\text{Or } -\dot{\omega} + a_x c - 2 \lambda (\omega - v c) + \lambda^2 (-\theta + \theta_{max}) \ge 0 \\
\end{split}
\end{equation}

\subsection{Probabilistic Control Barrier Functions}

If the state $X$ is uncertain, we can use probabilistic control barrier functions which ensure the satisfaction of the constraints by a confidence margin $\eta$ \citep{Lyu2021ProbabilisticSA}, i.e.,

\begin{equation}
\small
\begin{split}
    &Pr(\dddot{h}_{left}(X,U) + 3 \lambda \ddot{h}_{left}(X,U) + 3 \lambda^2 \dot{h}_{left}(X,U) +\\
    &\lambda^3 h_{left}(X,U) \geq 0) \geq \eta \\
    &Pr(\dddot{h}_{right}(X,U) + 3 \lambda \ddot{h}_{right}(X,U) + 3 \lambda^2 \dot{h}_{right}(X,U) +\\
    &\lambda^3 h_{right}(X,U) \geq 0) \geq \eta \\
    &Pr(\ddot{h}_{\theta,left}(X,U) + 2 \lambda \dot{h}_{\theta,left}(X,U) + \lambda^2 h_{\theta,left}(X,U) \ge 0) \geq \eta \\
    &Pr(\ddot{h}_{\theta,right}(X,U) + 2 \lambda \dot{h}_{\theta,right}(X,U) + \lambda^2 h_{\theta,right}(X,U) \ge 0) \geq \eta \\
    \end{split}
\end{equation}
If we represent the equation in the form:
    \begin{equation}
    \begin{split}
    &a^TU \leq b \text{ where } a \sim \mathbb{N} (\bar{a},\sigma_a) \\
    &Pr(a^Tc \leq b) = \phi\left(\frac{b-\bar{a}^T c}{\sqrt{c^T\Sigma c}}\right)\\
    &Pr(a^Tc \leq b) <=> b - \bar{a}^T u \geq \phi^{-1} (\eta)||\sigma_a^{1/2}c||^2
\end{split}
\end{equation}
In order to derive hard constraints on the control $u$ we conduct $n$ rollouts on the function $a(x)$ with different values of $x$ chosen at random and obtain the mean $\bar{a}$ and variance $\Sigma$. We fit a gaussian distribution to these, with the values $\bar{a}$ and $\sigma_a$ obtained as: $ \bar{a} = \frac{\Sigma_{i=0,1..n-1} a(X_i)}{n}$ and $\sigma_a = \sqrt{\frac{\Sigma_{i=0,1..n-1} (a(X_i)-\bar{a})}{n}}$, where $X_i$ is the state obtained from the $i^{th}$ rollout of the state obtained from the trained network and sensor inputs.


\subsection{Vehicle Models} \label{sec:veh_models}

\begin{figure}
    \centering
    \includegraphics[width=0.35\textwidth]{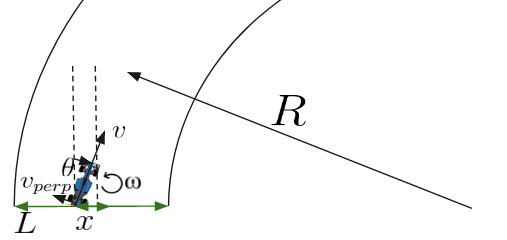}
    \caption{Vehicle state representation}
    \label{fig:state_rep}
\end{figure}

We use two different vehicle models. The complete state of the vehicle is illustrated in Fig. \ref{fig:state_rep}

\subsubsection{Kinematic model}

The kinematic bicycle model assumes no tire slip at low speed. Here, for the Frenet frame, as we only care about the lateral displacement of the vehicle, we define the state as $X = [x \ \theta \ v \ c]^T$ and the model is defined as follows: 

\begin{equation} \label{eq:dynamic_eqn}
\begin{bmatrix}
    \dot{x} \\
    \dot{\theta} \\
    \dot{v} \\
    \dot{c}
\end{bmatrix}
= \begin{bmatrix}
    v \sin \theta \\
    \frac{v \sin \delta}{L} - v c\\
    a_x \\
    0
\end{bmatrix}    
\end{equation}

\subsubsection{Dynamic model}

For high speeds, the kinematic model fails to accurately model the vehicle dynamics. To model the considerable amount of lateral slip, which is critical in racing scenarios, we use the linear tire model \citep{book} as follows where we take the state $X = [x \ \theta \ \omega \ v \ v_{perp} \ c]^T$.

\begin{equation}
\begin{bmatrix}
    \dot{x} \\
    \dot{\theta} \\
    \dot{\omega} \\
    \dot{v} \\
    \dot{v_{perp}} \\
    \dot{c}
\end{bmatrix}
= \begin{bmatrix}
    v \sin \theta + v_{perp} \cos \theta\\
    \omega - v c\\
    \frac{-2 (l_f^2 C_f + l_r^2 C_r)}{Iz v} \omega+ 2 \frac{l_f C_f}{Iz} \delta \\
    a_x \\
    \frac{-2 (C_f+C_r)}{m v} v_{perp} \\
    0
\end{bmatrix} 
\end{equation}



\subsection{Accounting for uncertainty in DNN predictions}

As we use a neural network to predict a subset of the state $X$, i.e., $[x \ \theta \ c]^T$, there are 2 types of uncertainties associated with the prediction from the trained network on a given dataset $D$ \citep{Hllermeier2021AleatoricAE}: aleatoric and epistemic. Aleatoric uncertainty is caused when the dataset is not accurate, i.e., for a single input, there is a distribution of outputs in the dataset and not a deterministic output. In other words, there are data elements with the same input but different outputs. In our case, this uncertainty would exist for the control prediction model if the expert driver strategy is stochastic. Epistemic uncertainty is the uncertainty caused due to the test input being outside the training distribution, i.e., the model is uncertain of the prediction if the input is unseen. In our case, the more prominent uncertainty is the epistemic one, as we use the MPC for expert feedback which is deterministic w.r.t. the given state and expect more uncertainty in the prediction if the input test image is outside the training data distribution. We use a common scheme used in the literature to estimate this epistemic uncertainty \cite{Gal2016DropoutAA}, which uses dropout layers in the network and then takes the mean $X_\mu$ and variance $X_{\sigma}$ (assuming all state variables to be independent) of the predictions from $n$ different rollouts of the model.   

\begin{equation}
\begin{split}
    &X_\mu = \frac{\Sigma_{i=0...n-1}F_i(X_{im})}{n} \\
    &X_\sigma = \sqrt{\frac{\Sigma_{i=0...n-1}(F_i(X_{im})-X_\mu)^2}{n}} \\
\end{split}
\end{equation}




\section{Training cycle} \label{sec:tr_cycle}

Putting both the imitation learning framework and the safety controller together, the process and training and testing are illustrated in Fig. \ref{fig:train_cycle}. 1) An MPC and full access to sensors are leveraged to control the vehicle in the first lap (iter 0) to collect data. 2) The data are used to train the end-to-end control neural network $M_{com}$ (mapping from image to control action) and the end-to-end state neural network $M_{st}$ (mapping from image to state). 3) In the following training, a safe controller using a quadratic programming-CBF (QP-CBF) is used to control the vehicle. The state for the QP-CBF is from the output of $M_{st}$. The MPC does not control the vehicle but provides control actions as labels for continuously training $M_{com}$. Meanwhile, $M_{st}$ is trained using the ground truth states from the sensors. 4) In the testing part, we directly use the two neural networks to obtaing the QP-CBF formulation to control the vehicle.

The QP-CBF is formulated as follows: 

\begin{figure}
    \centering
    \includegraphics[width=0.4\textwidth]{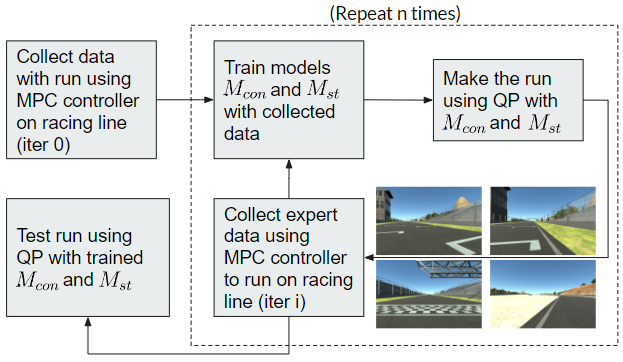}
    \caption{Training and testing in Carla. $M_{com}$: end-to-end control network; $M_{st}$: end-to-end state network}
    \label{fig:train_cycle}
\end{figure}

\vspace{-5mm}
\begin{equation} \label{obj_func_final}
\small
\begin{split}
    \mathop{\min}\limits_{U,\epsilon_i,\epsilon_2,\epsilon_3,\epsilon_4}&(U-U_{ref,\mu})^T U_\sigma^{-1} (U-U_{ref,\mu}) + \sum_{i=0}^{3} k_{err} (\epsilon_i^2)\\
    s.t.~~&Pr(\dddot{h}_{left}(X,U) + 3 \ddot{h}_{left}(X,U) + 3 \dot{h}_{left}(X,U) +\\ &h_{left}(X,U) \geq -\epsilon_1) \geq \eta \\
    &Pr(\dddot{h}_{right}(X,U) + 3 \lambda \ddot{h}_{right}(X,U) + 3 \lambda^2 \dot{h}_{right}(X,U)\\
    &+ \lambda^3 h_{right}(X,U) \geq -\epsilon_2) \geq \eta \\
    &Pr(\ddot{h}_{\theta,left}(X,U) + 2 \lambda \dot{h}_{\theta,left}(X,U) \\
    &+ \lambda^2 h_{\theta,left}(X,U) \ge -\epsilon_3) \geq \eta \\
    &Pr(\ddot{h}_{\theta,right}(X,U) + 2 \lambda \dot{h}_{\theta,right}(X,U) \\
    &+ \lambda^2 h_{\theta,right}(X,U) \ge -\epsilon_4) \geq \eta \\
\end{split}
\end{equation}

where $\epsilon_1$, $\epsilon_2$ are added to impose soft constraints on the lane violation. Constant $k$ dictates how severely to penalize if the constraints are violated. $U_{ref,\mu}$ and $U_\sigma$ are the mean and variance obtained from the control prediction obtained by $n$ rollouts on $M_{con}$. There are two things to note here: 1) When $M_{con}$ is uncertain, the magnitude of $U_\sigma$ is large, hence when the vehicle is at the boundary, the violation of lane constraints is given more priority, which drives the vehicle away from the boundary; 2) When $M_{con}$ is more certain and there is more uncertainty of the state from $M_{st}$, more weight is given to following the reference state $U_{ref,\mu}$ obtained from $M_{con}$.

\section{Experimental results}

We perform experiments on both the Carla simulator and an RC car. A video link to all the experiment recordings is available online \footnote[1]{https://www.youtube.com/watch?v=Ora8JCjprw8}. 

\subsection{Carla simulator}

For the Carla simulator experiments, we use the position obtained from the simulator and pre-labeled track boundaries to obtain the expert MPC controller. 

\subsubsection{Track}

\begin{figure}[htbp]
\begin{subfigure}{.21\textwidth}
    \centering
    \includegraphics[width=\textwidth]{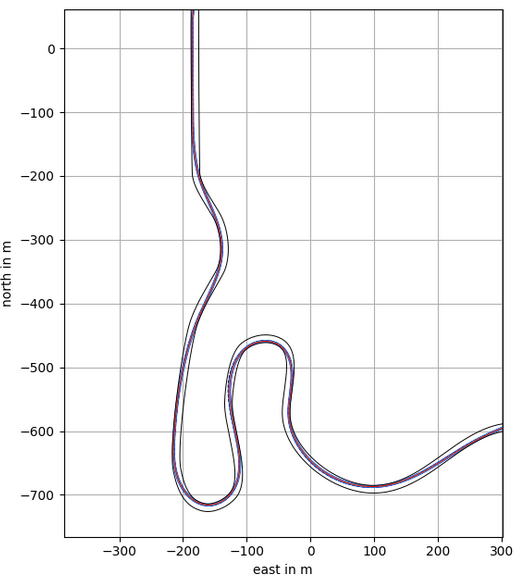}
    \caption{}
    \label{fig:racing_line}
\end{subfigure}
\begin{subfigure}{.27\textwidth}
    \centering
    \includegraphics[width=\textwidth]{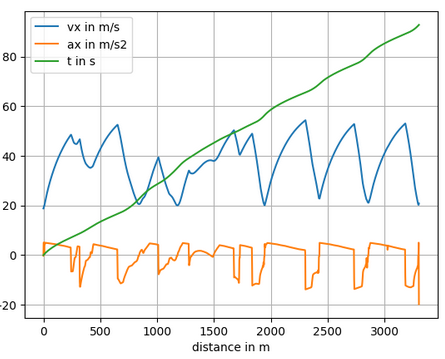}
    \caption{}
    \label{fig:racing_line_speeds}
\end{subfigure}
\caption{Racing line}
\label{fig:racing_line_}
\end{figure}

Our test track is shown in Fig. \ref{fig:racing_line} 
We use only about 70\% of the track for training and the rest for testing. The optimal racing line is shown in Fig. \ref{fig:racing_line} with velocity profile in Fig. \ref{fig:racing_line_speeds}.


\subsubsection{Vehicle model and Hyperparameter settings}

We use a dynamic vehicle model for the expert controller and CBF constraint due to high speed. We use $\lambda=2.5$, $\theta_{max} = \frac{\pi}{4}$, $\eta=0.95$ 

\subsubsection{Using center line as reference}

\begin{figure}[htbp]
\begin{subfigure}{.24\textwidth}
    \centering
    \includegraphics[width=\textwidth]{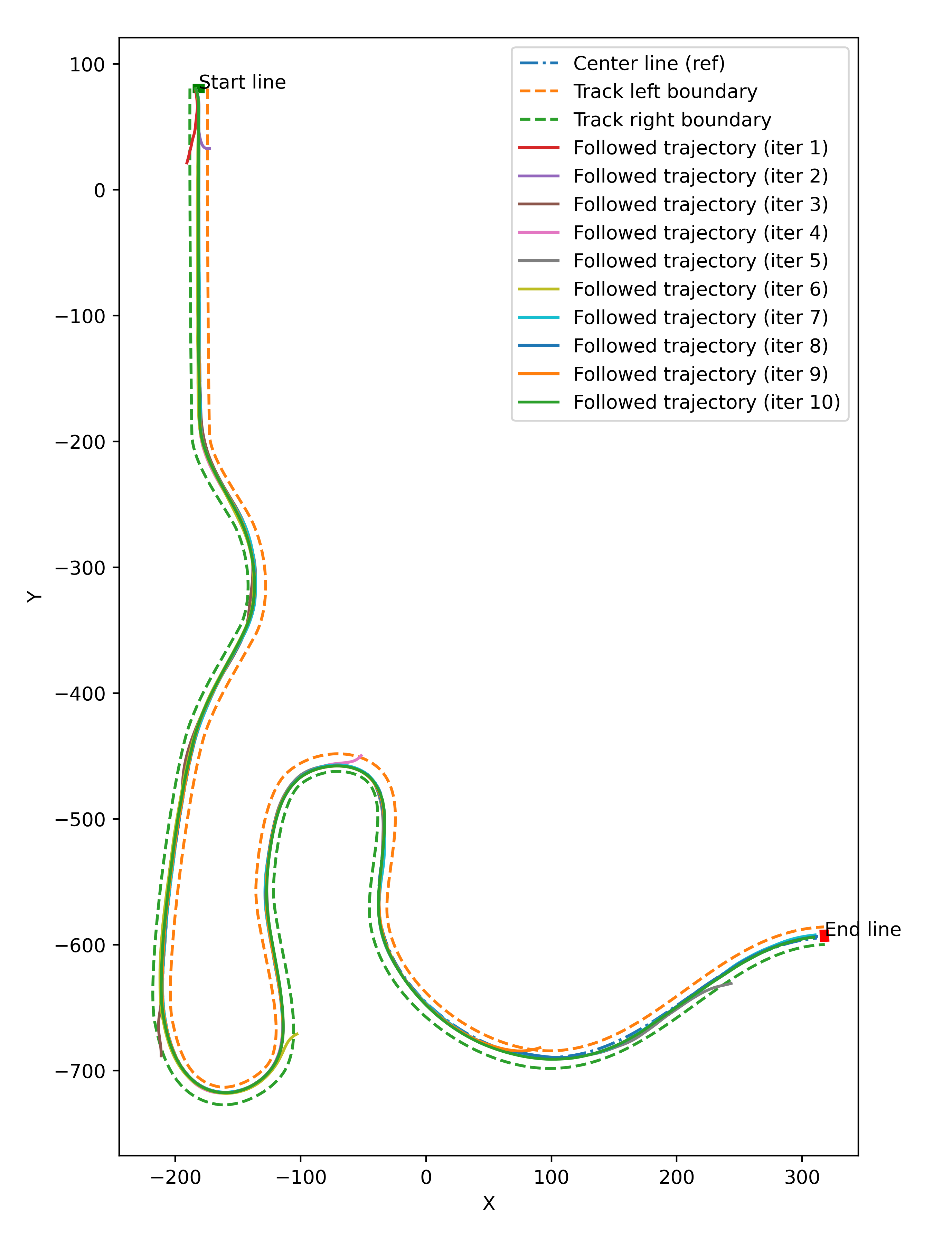}
    \caption{Without CBF}
    \label{fig:cent_line_without_cbf}
\end{subfigure}
\begin{subfigure}{.24\textwidth}
    \centering
    \includegraphics[width=\textwidth]{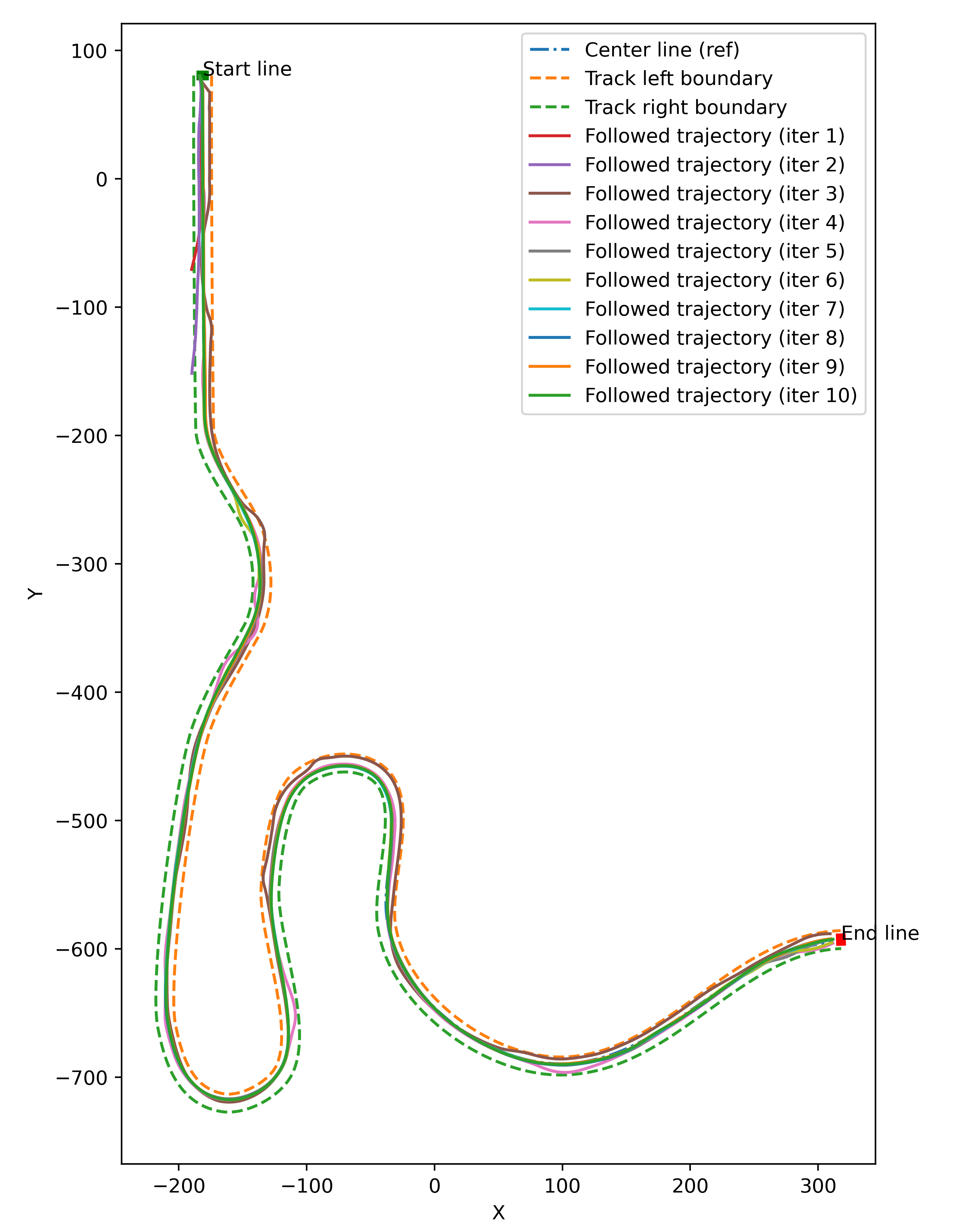}
    \caption{With CBF}
    \label{fig:cent_line_with_cbf}
\end{subfigure}
\caption{Trajectories for all iterations (Carla, Center line)}
\label{fig:cent_line}
\end{figure}

First, we train an end-to-end policy to run on the center line of the track. The results without CBF and with CBF are shown in Fig. \ref{fig:cent_line_without_cbf} and Fig. \ref{fig:cent_line_with_cbf}, respectively. As can be observed, with the safety controller, the vehicle learns to complete the track in less time and gets reasonably good lap times, which improve with each iteration, as shown in Table \ref{table:race_times_cent_line}.

\begin{table}[htbp]
\begin{center}
\begin{tabular}{| m{0.4cm} | m{0.6cm} | m{1.1cm} | m{1.3cm} | m{1.1cm} | m{1.3cm} |} 
\hline
\multirow{2}{*}{Iter} & \multirow{2}{*}{CBF?} & \multicolumn{2}{c|}{Ref : Center line (k=2)} & \multicolumn{2}{c|}{Ref : Racing line (k=4)} \\
\cline{3-6}
 &  & Lap time (in s) & Mean deviation from ref & Lap time (in s) & Mean deviation from ref \\ [0.5ex] 
\hline
 0 & $\times$ & 70.9$^*$ & 0.05 & 59.4 & 0.1 \\ 
 \hline
 1 & $\times$ & 10.1$^*$ & 2.94$^*$ & 5.7$^*$ & {3.62$^*$} \\ 
 \hline
 2 & $\times$ &  {9.5}$^*$ &  {0.96}$^*$ &  {4.65}$^*$ &  {4.63}$^*$ \\ 
 \hline
 \multirow{2}{*}{3} & $\times$ &  {37.0}$^*$ &  {0.91}$^*$ &  {11.55}$^*$ &  {1.76}$^*$ \\ 
 \cline{2-6}
  & $\checkmark$ & 72.8 & 3.49 & - & - \\ 
 \hline
 \multirow{2}{*}{4} & $\times$ &  {54.4}$^*$ &  {0.54}$^*$ &  {14.1}$^*$ &  {1.81} \\ 
 \cline{2-6}
 & $\checkmark$ & 71.6 & 1.19 & - & - \\
 \hline
\multirow{2}{*}{5} & $\times$ &  {68.9}$^*$ &  {0.61}$^*$ &  {42.0}$^*$ &  {3.62}$^*$ \\ 
 \cline{2-6}
 & $\checkmark$ & 70.9 & 0.54 &  {24.75}$^*$ &  {1.49}$^*$\\
 \hline
 \multirow{2}{*}{6} & $\times$ &  {43.4}$^*$ &  {1.17}$^*$ &  {19.65}$^*$ &  {2.08}$^*$ \\ 
 \cline{2-6}
 & $\checkmark$ & 70.9 & 0.35 & 60.6 & 2.17 \\
 \hline
 \multirow{2}{*}{7} & $\times$ & 71.0 & 0.39 &  {37.2}$^*$ &  {0.65}$^*$ \\ 
 \cline{2-6}
 & $\checkmark$ & 70.8 & 0.26 & 59.85 & 1.386 \\
 \hline
 \multirow{2}{*}{8} & $\times$ & 70.9 & 0.35 &  {18.45}$^*$ &  {1.99}$^*$ \\ 
 \cline{2-6}
 & $\checkmark$ & 70.9 & 0.27 & 59.85 & 1.01 \\
 \hline
 \multirow{2}{*}{9} & $\times$ & - & - &  {40.35}$^*$ &  {0.40}$^*$ \\ 
 \cline{2-6}
 & $\checkmark$ & - & - & 59.7 & 1.22 \\
 \hline
 \multirow{2}{*}{10} & $\times$ & - & - & 60.55 & 1.41 \\ 
 \cline{2-6}
 & $\checkmark$ & - & - & 59.7 & 0.741 \\
 \hline
 \multirow{2}{*}{11} & $\times$ & - & - & 59.8 & 1.16 \\ 
 \cline{2-6}
 & $\checkmark$ & - & - & 59.7 & 0.437 \\
 \hline
\end{tabular}
\end{center}
\caption{Statistics for all iterations (Carla simulator). $^*$ : Didn't complete the track}
\label{table:race_times_cent_line}
\end{table}

\subsubsection{Using racing line as reference}

We obtain the results of using the racing line as the reference. As expected, as it is more challenging to learn to run with the racing line as reference, without using the safety CBF, the end-to-end policy learns to complete the track in about $11$ iterations, whereas using CBF, it completes the track in only $5$ iterations and with improved lap times later. The lap times are shown in Table \ref{table:race_times_cent_line}. Hence, we can expect that using the safety CBF safely drives the vehicle along the track with fewer iterations required for training.

\begin{figure}[htbp]
\begin{subfigure}{.23\textwidth}
    \centering
    \includegraphics[width=\textwidth]{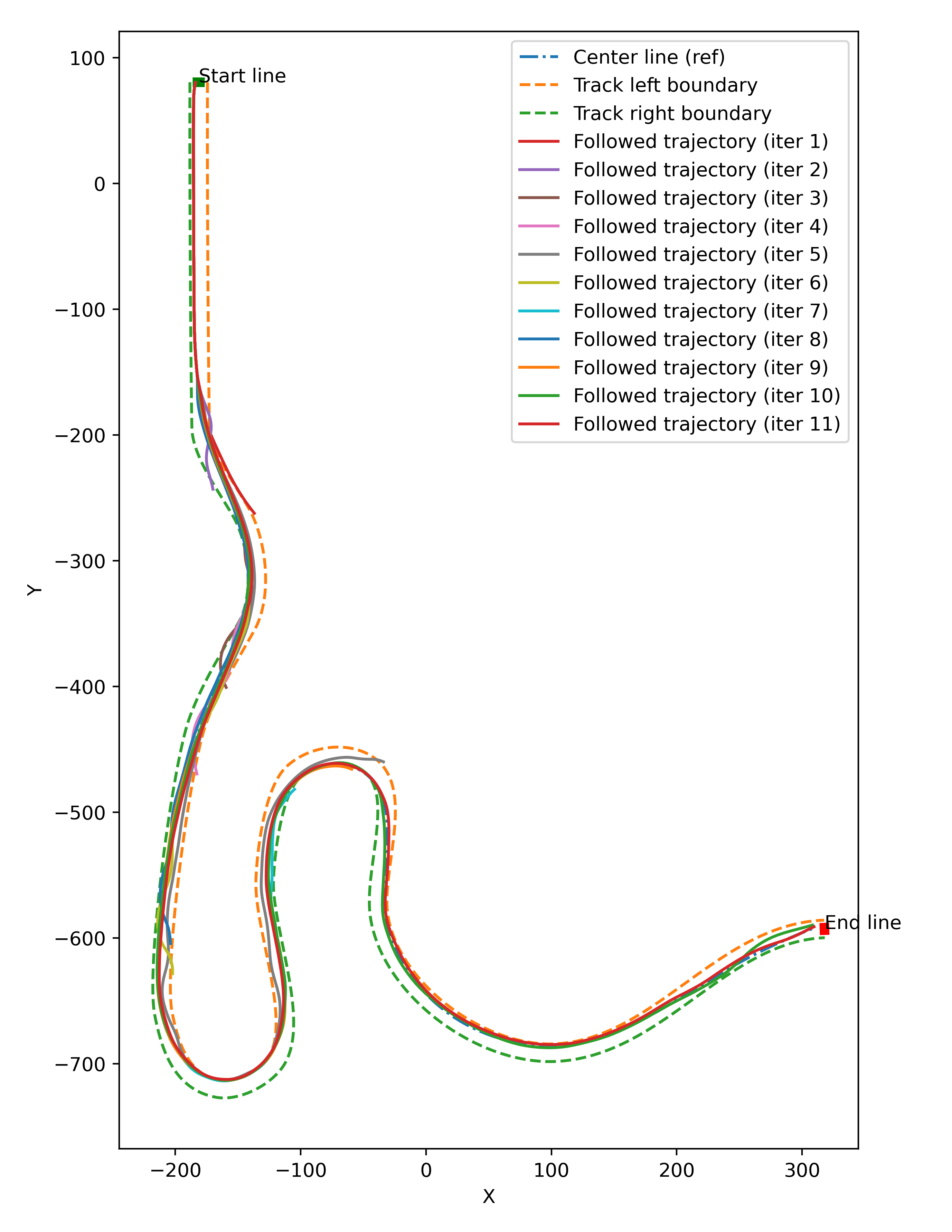}
    \caption{Without CBF}
    \label{fig:race_line_without_cbf}
\end{subfigure}
\begin{subfigure}{.23\textwidth}
    \centering
    \includegraphics[width=\textwidth]{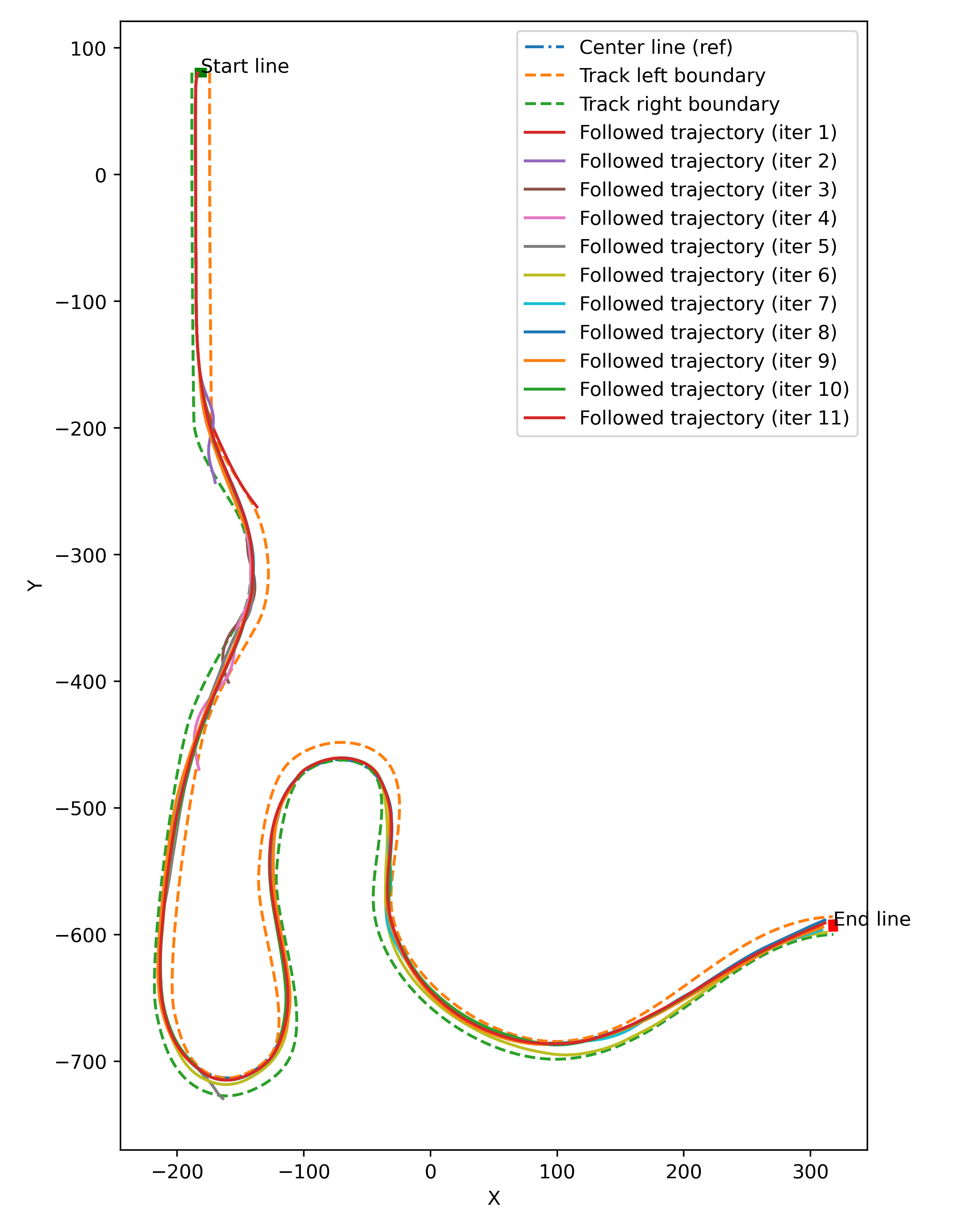}
    \caption{With CBF}
    \label{fig:race_line_with_cbf}
\end{subfigure}
\caption{Trajectories for all iterations (Carla, Racing line)}
\label{fig:race_line1}
\end{figure}



\subsection{RC car}

Finally, we also perform the experiments on a real $1/10$-scale RC car which is fitted with Lidar, depth dual camera, IMU and wheel encoders. It has an NVIDIA Jetson TX2 attached to it with 8GB of RAM and 256 core NVIDIA Pascal GPU for onboard computation. There is a twin ZED camera attached to the vehicle which also outputs depth information. However we only take the first 3 channels of the input excluding the depth information of the camera. Also, as there are two camera image feeds obtained from the twin camera, we merge both the images to obtain a $6$-channel image and pass it as input to the state and control prediction networks. 

\begin{figure}[htbp]
\begin{subfigure}{.281\textwidth}
    \centering
    \includegraphics[width=\textwidth]{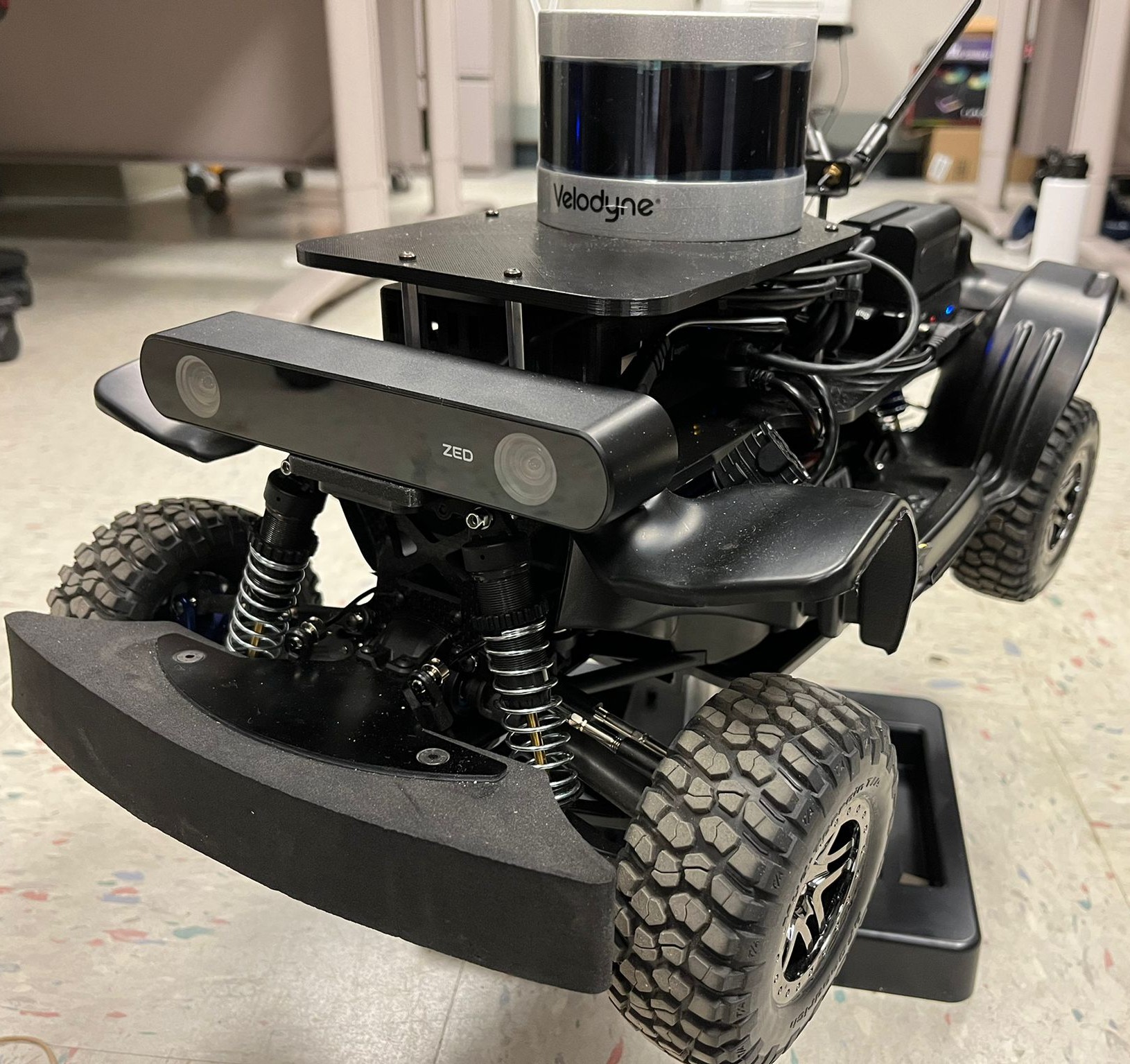}
    \caption{Hardware}
    \label{fig:rc_car_look}
\end{subfigure}
\begin{subfigure}{.2\textwidth}
    \centering
    \includegraphics[width=\textwidth]{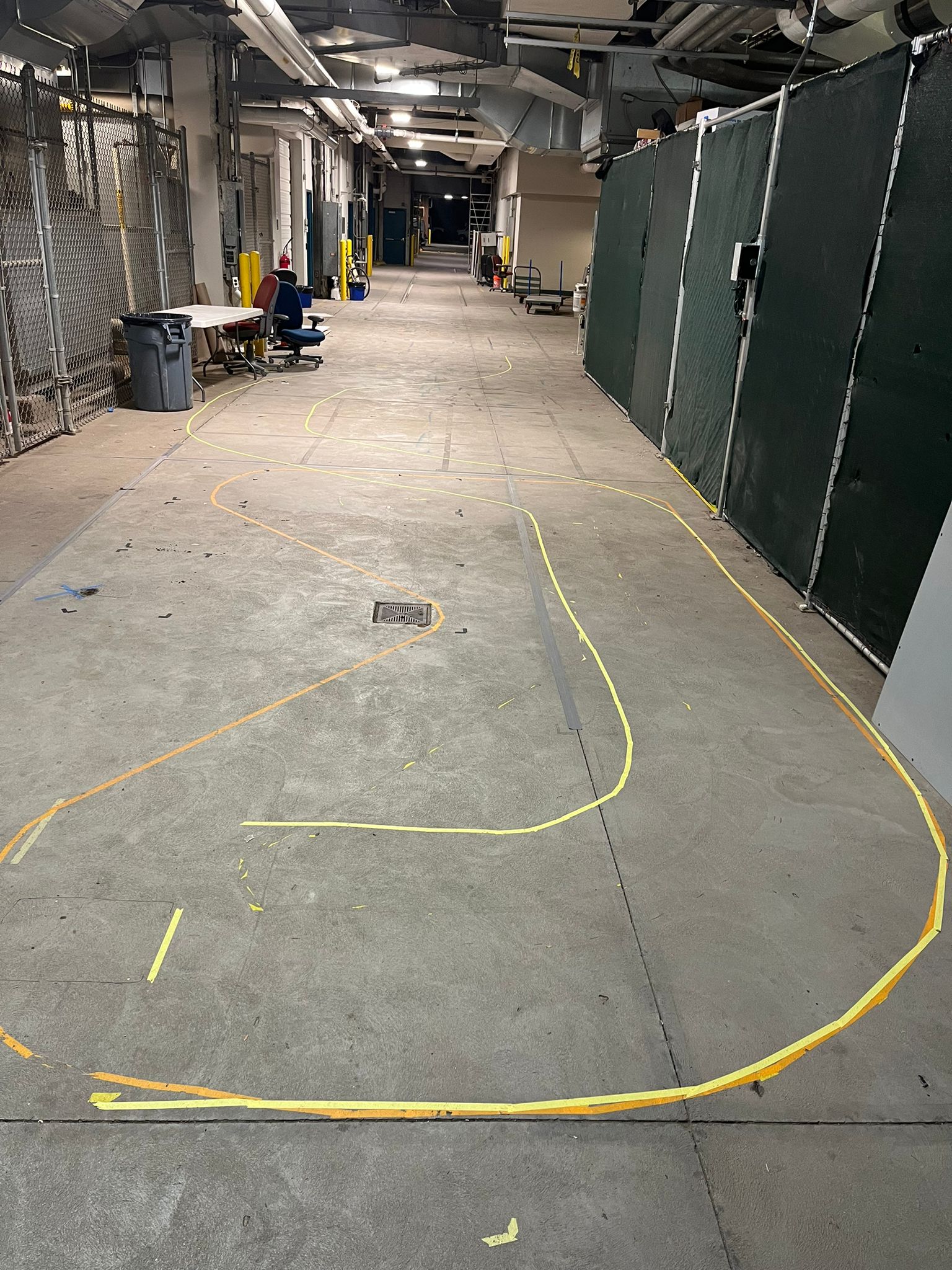}
    \caption{Track}
    \label{fig:rc_car_track}
\end{subfigure}
\caption{RC car setup}
\label{fig:rc_car}
\end{figure}

\subsubsection{Track}

\begin{figure}[htbp]
\begin{subfigure}{.19\textwidth}
    \centering
    \includegraphics[width=\textwidth]{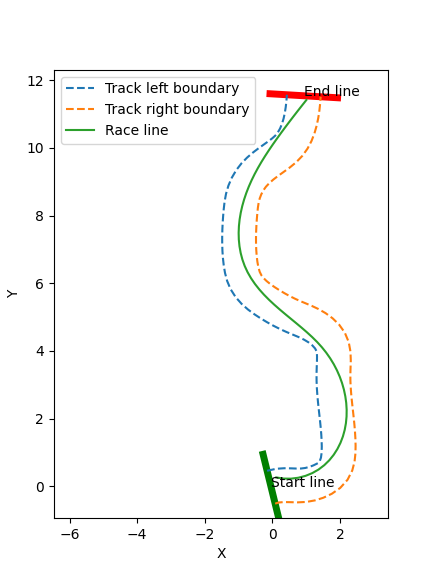}
    \caption{}
    \label{fig:racing_line_rc}
\end{subfigure}
\begin{subfigure}{.29\textwidth}
    \centering
    \includegraphics[width=\textwidth]{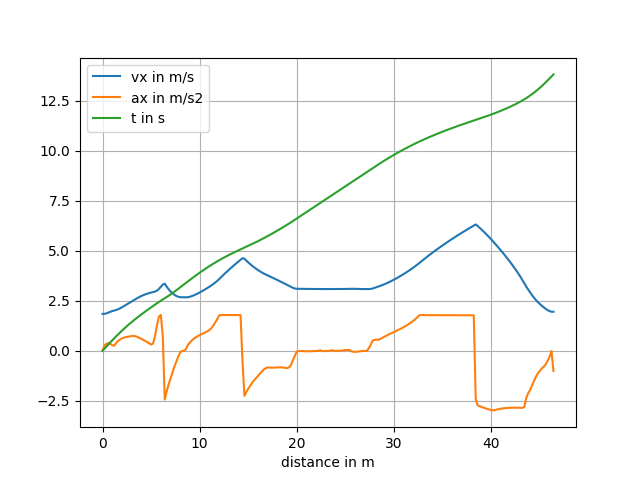}
    \caption{}
    \label{fig:racing_line_speeds_rc}
\end{subfigure}
\caption{Racing line}
\label{fig:racing_line_rc_}
\end{figure}

We perform the experiment on the track shown in Fig. \ref{fig:racing_line_rc}. The lane width is about 0.8 m. The racing line and the reference speeds are shown in Fig. \ref{fig:racing_line_rc} and Fig. \ref{fig:racing_line_speeds_rc}, respectively.

\subsubsection{Localization for training}

We use hector SLAM \citep{Saat2019HECTORSLAM2M} for pre-mapping and Monte Carlo localization (MCL) for localization \citep{Dellaert1999MonteCL}.

\subsubsection{Vehicle model}

We use the kinematic vehicle model for the expert controller and CBF due to the low-speed capacity of the vehicle.

\subsubsection{Using center line as reference}

\begin{figure}[htbp]
\begin{subfigure}{.249\textwidth}
    \centering
    \includegraphics[width=\textwidth]{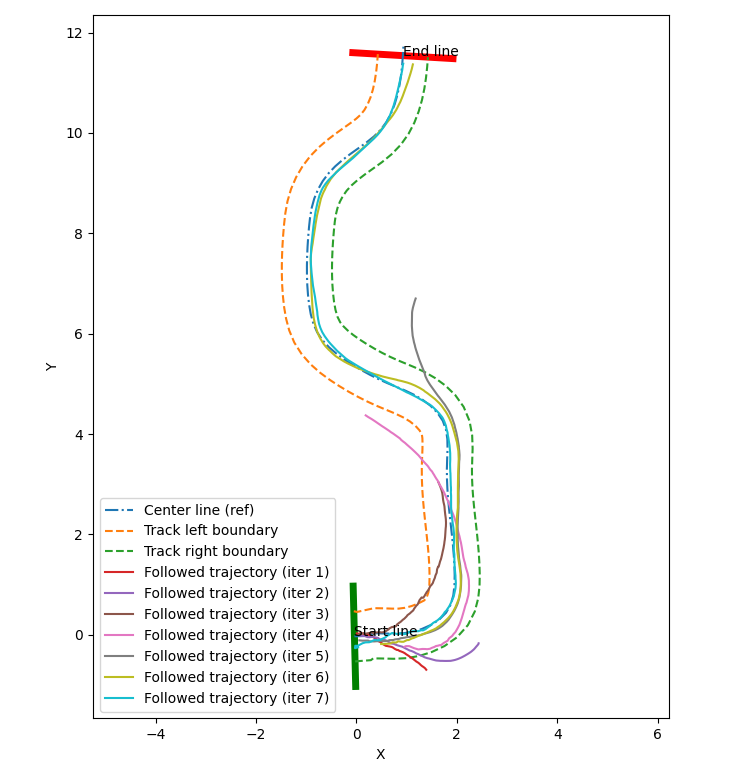}
     \caption{Without CBF}
    \label{fig:cent_line_without_cbf_rc}
\end{subfigure}
\begin{subfigure}{.231\textwidth}
    \centering
    \includegraphics[width=\textwidth]{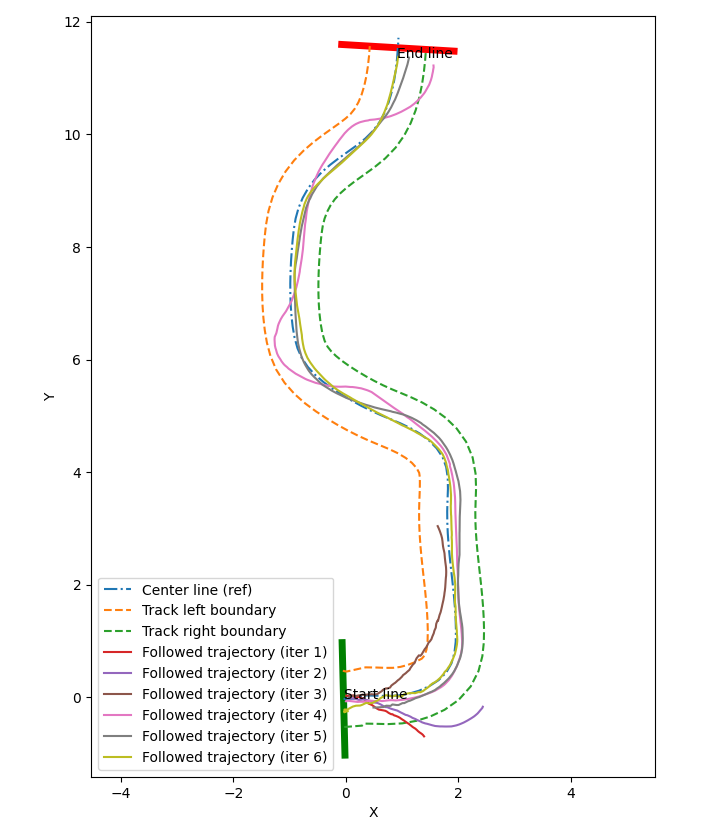}
    \caption{With CBF}
    \label{fig:cent_line_with_cbf_rc}
\end{subfigure}
\caption{Trajectories for all iterations (RC car, Center line)}
\label{fig:cent_line_rc}
\end{figure}

Using the center line as the reference, the results are shown in Table \ref{table:race_times_cent_line_rc} and Figures \ref{fig:cent_line_without_cbf_rc} and \ref{fig:cent_line_with_cbf_rc}. We observe a similar trend to that of the simulator experiments.

\begin{table}[htbp]
\begin{center}
\begin{tabular}{| m{0.4cm} | m{0.6cm} | m{1.3cm} | m{1.1cm} | m{1.3cm} | m{1.1cm} |} 
\hline
\multirow{2}{*}{Iter} & \multirow{2}{*}{CBF?} & \multicolumn{2}{c|}{Ref : Center line (k=3)} & \multicolumn{2}{c|}{Ref : Racing line (k=3)} \\
\cline{3-6}
 &  & Mean deviation from ref & Lap time (in s)  & Mean deviation from ref & Lap time (in s) \\ [0.5ex] 
\hline
 0 & $\times$ & 0.02 & 30.98 & 0.024 & 20.73 \\ 
 \hline
 1 & $\times$ &  {0.328}$^*$ &  {5.13}$^*$ & 0.341$^*$ & 2.41$^*$ \\ 
 \hline
 2 & $\times$ &  {0.509}$^*$ &  {5.78}$^*$ & 0.217$^*$ & 3.24$^*$ \\ 
 \hline
 3 & $\times$ &  {0.210}$^*$ &  {8.77}$^*$ & 0.223$^*$ & 9.26$^*$ \\ 
 \hline
 \multirow{2}{*}{4} & $\times$ &  {0.379}$^*$ &  {16.56}$^*$ & 0.278$^*$ & 11.85$^*$ \\ 
 \cline{2-6}
 & $\checkmark$ & 0.175 & 33.63 & 0.334 & 24.91 \\
 \hline
\multirow{2}{*}{5} & $\times$ &  {0.340}$^*$ &  {18.58}$^*$ & 0.260$^*$ & 12.31$^*$ \\ 
 \cline{2-6}
 & $\checkmark$ & 0.092 & 31.15 & 0.277 & 24.80 \\
 \hline
 \multirow{2}{*}{6} & $\times$ & 0.110 & 31.23 & 0.223$^*$ & 12.9$^*$ \\ 
 \cline{2-6}
 & $\checkmark$ & 0.048 & 31.03 & 0.201 & 24.91 \\
 \hline
 \multirow{2}{*}{7} & $\times$ & 0.048 & 30.96 &  0.141$^*$ & 14.92$^*$ \\ 
 \cline{2-6}
 & $\checkmark$ & 0.044 & 31.17 & 0.291 & 23.69 \\
 \hline
 \multirow{2}{*}{8} & $\times$ & - & - & 0.159$^*$ & 15.62$^*$ \\ 
 \cline{2-6}
 & $\checkmark$ & - & - & 0.223 & 23.47 \\
 \hline
 \multirow{2}{*}{9} & $\times$ & - & - & 0.291 & 23.68 \\ 
 \cline{2-6}
 & $\checkmark$ & - & - & 0.113 & 23.28 \\
 \hline
\end{tabular}
\end{center}
\caption{Statistics for all iterations (RC car). $^*$: Didn't complete the track}
\label{table:race_times_cent_line_rc}
\end{table}

\subsubsection{Using racing line as reference}

\begin{figure}[htbp]
\begin{subfigure}{.24\textwidth}
    \centering
    \includegraphics[width=\textwidth]{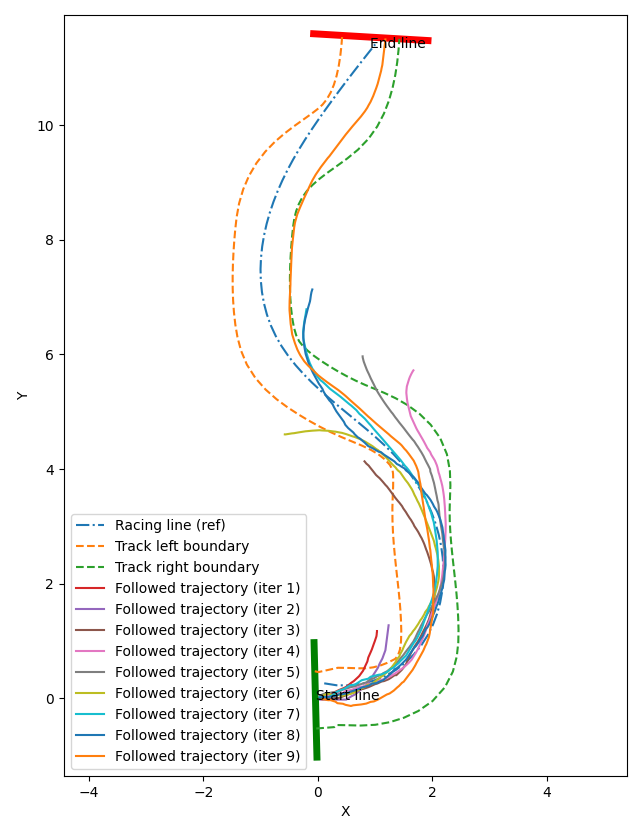}
    \caption{Without CBF}
    \label{fig:race_line_without_cbf_rc}
\end{subfigure}
\begin{subfigure}{.24\textwidth}
    \centering
    \includegraphics[width=\textwidth]{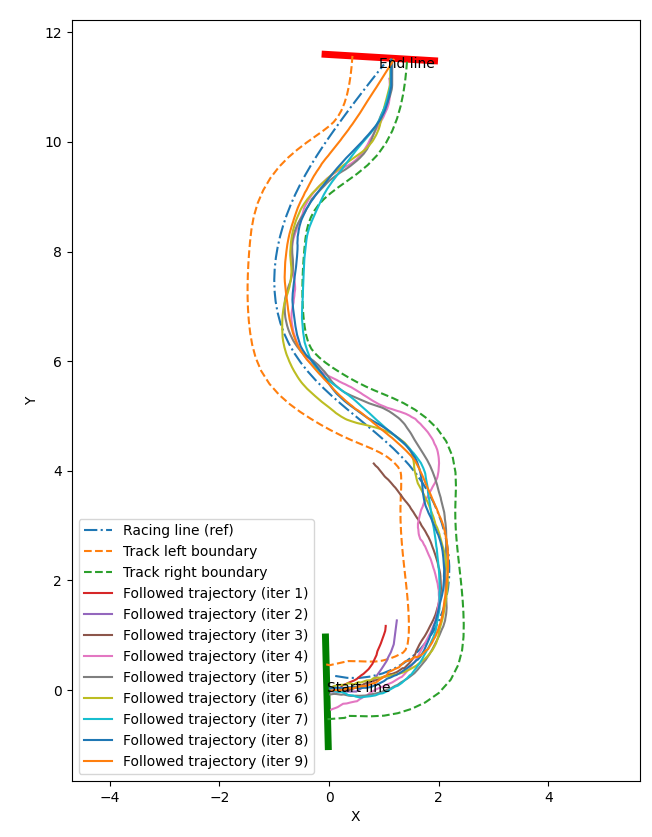}
    \caption{With CBF}
    \label{fig:race_line_with_cbf_rc}
\end{subfigure}
\caption{Trajectories for all iterations (RC car, Racing line)}
\label{fig:race_line1_rc}
\end{figure}

Finally, the results using the racing line as the reference are tabulated in Table \ref{table:race_times_cent_line_rc} and Figures \ref{fig:race_line_without_cbf_rc} and \ref{fig:race_line_with_cbf_rc}. We observe a similar trend to that of the simulator experiments.

\subsubsection{Visualizing feature maps}

\begin{figure}[htbp]
\begin{subfigure}{.24\textwidth}
    \centering
    \includegraphics[width=\textwidth]{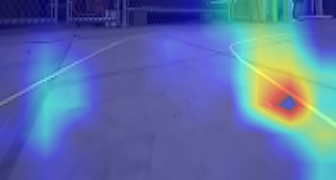}
    \caption{Left}
\end{subfigure}
\begin{subfigure}{.24\textwidth}
    \centering
    \includegraphics[width=\textwidth]{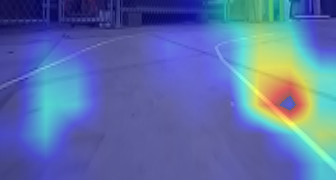}
    \caption{Right}
\end{subfigure}
\begin{subfigure}{.24\textwidth}
    \centering
    \includegraphics[width=\textwidth]{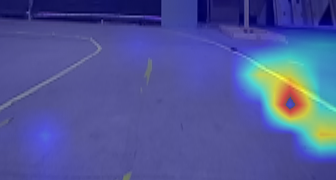}
    \caption{Left}
\end{subfigure}
\begin{subfigure}{.24\textwidth}
    \centering
    \includegraphics[width=\textwidth]{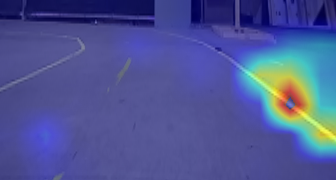}
    \caption{Right}
\end{subfigure}
\caption{GradCAM projections on two example images}
\label{fig:gradcam}
\end{figure}

We also visualize what parts of the image maximally contribute to the steering and speed predictions in the trained network of the RC car (See Fig. \ref{fig:gradcam}) using GradCAM \citep{Selvaraju2017GradCAMVE}. GradCAM is a popular tool used in  the literature to visualize which part of the image contributes  the maximum to the class label predicted by the model. In the case of a regression model, we use the reduction function as the identity of the output value, as discussed in \cite{Wang2018DiabeticRD}. This would mean finding the region of the image contributing the maximum to the regression output, which is steering and speed in our case. As can be observed from the figures, most of the contribution is made by the track boundary markings, which means that the model learns to derive its control by looking at the lane markings relative to the RC car. In our case, as we have two images stacked across channels as input to the network, we project the activation maps on both the left and right camera images for visualization in Fig. \ref{fig:gradcam}.

\section{Conclusion} \label{sec:conclusion}

This paper presents a novel algorithm that uses end-to-end DNN for controlling a racing vehicle and a safety controller to explicitly safeguard the commands predicted from the end-to-end model. The epistemic uncertainty in the predictions of states from DNN is also considered as part of the framework. In future work, it would be interesting to use lifelong learning frameworks for offline learning to not have to train the model on the whole dataset again but, only on the freshly collected data of the last run. Also extending the framework for online learning, using imitation learning-based RL on top of this, and transfer learning for track environment change would be an interesting avenue to explore in the future.


\bibliography{ifacconf}             
                                                   
\end{document}